\documentclass[letterpaper, 10 pt, conference]{ieeeconf}  
\usepackage[T1]{fontenc}
\usepackage{ae,aecompl}
\usepackage{amssymb}
\usepackage{amsmath}
\usepackage{amsmath,mathtools}
\usepackage{times}
\usepackage{caption}
\usepackage[labelformat=simple]{subcaption}
\usepackage{graphicx}
\usepackage{graphics}
\usepackage{blkarray}
\usepackage{booktabs}
\usepackage{xcolor} 
\usepackage{bbold}
\usepackage{caption}
\usepackage{subcaption}
\usepackage{multirow}
\usepackage{comment}
\usepackage{epstopdf}
\usepackage{tikz}
\usepackage{todonotes}
\graphicspath{{./figures/}}
\usepackage{csquotes}
\usepackage{siunitx}
\usepackage{cite}
\usepackage{epigraph}
\usepackage{empheq}
\usepackage{comment}
\usepackage{cancel}
\usepackage{algorithm} 
\usepackage{algorithmic}  
\usepackage{mathtools}

\usepackage[algo2e,linesnumbered,ruled,vlined]{algorithm2e}

\usepackage{hyperref}

\newcommand{\bt}{\mathcal{T}}

\SetKwRepeat{Do}{do}{while}

\makeatother

\DeclareCaptionLabelSeparator{periodspace}{.\quad}
\IEEEoverridecommandlockouts                              
\newtheorem{lemma}{Lemma}
\newtheorem{remark}{Remark}
\newtheorem{assumption}{Assumption}
\newtheorem{example}{Example}
\newtheorem{proposition}{Proposition}

\newtheorem{problem}{Problem} 

\newtheorem{definition}{Definition}
\newtheorem{experiment}{Case}

\makeatletter
\newcommand*{\rom}[1]{\expandafter\@slowromancap\romannumeral #1@}
\makeatother

\def\eqalignno#1{\let\\=\cr\displ@y \tabskip\@centering
  \halign to\displaywidth{\hfil$\@lign\displaystyle{##}$\tabskip\z@skip
    &$\@lign\displaystyle{{}##}$\hfil\tabskip\@centering
    &\llap{$\@lign##$}\tabskip\z@skip\crcr
    #1\crcr}}
\def\leqalignno#1{\let\\=\cr\displ@y \tabskip\@centering
  \halign to\displaywidth{\hfil$\@lign\displaystyle{##}$\tabskip\z@skip
    &$\@lign\displaystyle{{}##}$\hfil\tabskip\@centering
    &\kern-\displaywidth\rlap{$\@lign##$}\tabskip\displaywidth\crcr
    #1\crcr}}
\begin{document}

\thispagestyle{empty}
\twocolumn
\title{\LARGE \bf
Improving the Parallel Execution of Behavior Trees}

\author{Michele Colledanchise and Lorenzo Natale  
\thanks{The authors are with the iCub Facility, Istituto Italiano di Tecnologia. Genoa, Italy.
e-mail: {\tt{ michele.colledanchise@iit.it}} \newline This work was carried out in the context of the CARVE project, which has received funding from the European Union's Horizon 2020 research and innovation programme under grant agreement No 732410, in the form of financial support to third parties of the RobMoSys project.} }

\maketitle
\thispagestyle{empty}
\pagestyle{empty}

\begin{abstract}

Behavior Trees (BTs) have become a popular framework for designing controllers of autonomous agents in the computer game and in the robotics industry.
One of the key advantages of BTs lies in their modularity, where independent modules can be composed to create more complex ones. In the classical formulation of BTs, modules can be composed using one of the three operators: Sequence, Fallback, and Parallel. The Parallel operator is rarely used despite its strong potential against other control architectures such as Finite State Machines. This is due to the fact that concurrent actions may lead to unexpected problems similar to the ones experienced in concurrent programming.  
In this paper, we outline how to tackle the aforementioned problem by introducing Concurrent BTs (CBTs) as a generalization of BTs in which we include the notions of progress and resource usage. We show how CBTs allow safe concurrent executions of actions and we analyze the approach from a mathematical standpoint. 
To illustrate the use of CBTs, we provide a set of use cases in realistic robotics scenarios.

\end{abstract}
\section{Introduction}
\label{sec:introduction}

Behavior Trees (BTs) were introduced in the computer game industry to model the behavior of Non-Player Characters (NPCs)~\cite{isla2005handling,champandard2007understanding,isla2008halo}. BTs have now matured to the level that they are included in several textbooks on the topic~\cite{millington2009artificial,rabin2014gameAiPro} and game engines.
Following the development in computer game industry, BTs now also finds large use in real-robot applications, including manipulation~\cite{Bagnell2012b},
non-expert programming~\cite{guerin2015manufacturing}, 
 brain surgery \cite{hu2015ablation}. Other works include UAV missions  \cite{klockner2013,ogren}, the computer verification of plans~\cite{klockner2013}, and the estimation of resulting execution times in~\cite{Colledanchise14}.

BTs are appreciated because they are
modular, flexible, and reusable, and have also been shown
to generalize successful control architectures such
as the Subsumption architecture~\cite{brooks1986robust} and the Teleo-reactive
Paradigm~\cite{nilsson1994teleo} in~\cite{BTBook}.

The Parallel operator in BTs has the great advantage that it helps to tame the \emph{curse of dimensionality} that affects many other control architectures, in which the system's complexity is the product of its sub-systems' complexities~\cite{colledanchise2016advantages}. However, in the classical formulation of BTs, the trees in a Parallel operator are independent from each other. This limits their applications to orthogonal tasks~\cite{BTBook,champandard2007enabling}, making the Parallel operator rarely used compared to the other ones.

We propose Concurrent BTs (CBTs) as a generalization of BTs where for each node we define its progress value and the resources needed for its execution. We show how CBTs allow actions to be safely executed concurrently by adopting solutions from computer programming in terms of process synchronization and exclusive use of critical resources, as in the two following examples. 
\begin{figure}[t!]
\centering
\begin{subfigure}[b]{0.45\columnwidth}
\includegraphics[width=\columnwidth]{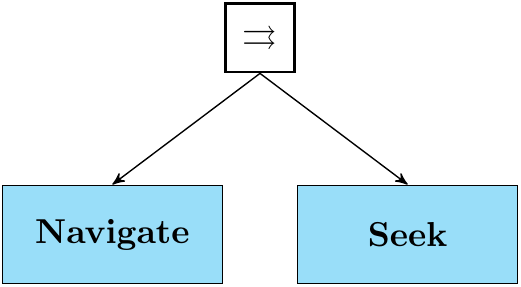}
\caption{Example of two BTs executed in parallel that require process synchronization.}
\label{intro.fig.sync}
\end{subfigure}
~
\begin{subfigure}[b]{0.45\columnwidth}
\includegraphics[width=\columnwidth]{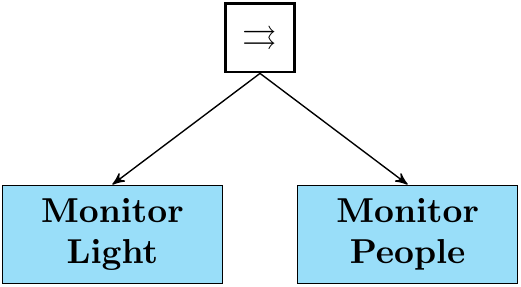}
\caption{Example of two BTs executed in parallel that require resource synchronization.}
\label{intro.fig.mutex}
\end{subfigure}
\caption{Example of two BTs executing tasks in parallel that require synchronization.}
\end{figure}

1) A robot that is asked to find and recognize the objects on the floor. A possible way to describe this task is using the BT in Figure~\ref{intro.fig.sync}. Where the action \emph{Navigate} generates and makes the robot follow a path and the action \emph{Seek} looks for objects on the floor and recognizes them. However, since classical formulation of BTs executes the two actions independently, there could be the case where the robot navigates in the room too fast for the seeking routine to recognize objects, resulting in an unsuccessful execution of the task. 

2) A service robot, beside its main task, has also to monitor the light condition of the room and whether a person gets too close to the robot. A possible way to describe this monitoring task is using the BT in Figure~\ref{intro.fig.mutex}. Each task has a warning message that invokes the speaker. However, the speaker allows one speech at a time. For this minor action, the tasks cannot be considered orthogonal and cannot be safely executed in parallel using the classical formulation of BTs. 

It is clear that the parallel execution of BTs needs synchronization strategies for exploiting its full potential. We believe that proper parallel task executions will bring benefits in terms of efficiency and multitasking to robot programming in a similar way as done in computer programming. However, in allowing arbitrary BTs to be executed in parallel we inherit fundamental problems of parallel computing: \emph{process synchronization}, where a process has to wait for another process for the execution, and \emph{data synchronization}, where the access to critical resources has to be regulated.

It is important to stress that execution of some routines may require different amount of time depending on the context (e.g. object recognition may be triggered only when visual inspection spots a potential object). Moreover there could be a substantial difference between the execution of an action in a simulation and in the real world. Therefore synchronization must be ensured at run-time.

The contributions of this paper are the definition of CBTs, in the Parallel composition of CBTs for process and resource synchronization, and a formal mathematical analysis. To the best of our knowledge, this is the first attempt at synchronizing the concurrent executions of BTs as it is done in computer programming.

The remainder of this paper is structured as follows: In Section~\ref{sec:related} we overview the related work. In Section~\ref{sec:background} we present the classical formulation of BTs. Then, in Section~\ref{sec:problem}, we formulate the problem and in Section~\ref{sec:proposed} we show the proposed solution. In Section~\ref{sec:implementation} some practical observations and in Section~\ref{sec:analysis} we analyze the solution from a mathematical standpoint. Finally, in Section~\ref{sec:results}, we present some use cases. We conclude the paper in Section~\ref{sec:conclusions}.

\section{Related Work}
\label{sec:related}

The Parallel operator is the least used operator in BT applications as it entails concurrency problems (e.g. race conditions, starvation, deadlocks, etc). Their applications usually assume one of the following: 1) the actions executed in parallel lie on orthogonal state spaces \cite{champandard2007enabling} or 2) the actions that can be potentially executed in parallel have a predefined priority assigned~\cite{weber2010reactive}.

In~\cite{ColledanchiseTRO17} the authors use a Parallel operator to monitor user requests and execute an activity accordingly. Since the user can define one activity at a time, the activities are mutually exclusive. Hence the BT for each activity (and its actions) is defined in its own sub-space.

In~\cite{colledanchise2016advantages} the authors show how to improve fault tolerance
and performance of a single robot
BT by adding more robots and extending the
BT into a so-called \emph{multirobot} BT. The authors show that, under some assumptions, by turning Sequence and Fallback operators of the original BT into a Parallel we can improve task performance.
The Parallel operator involves multiple agents, each assigned to a specific task by a task-assignment algorithm. The task-assignment algorithm ensures the absence of conflicts.

A more recent work~\cite{2018arXiv180107864H} uses BTs to represent medical procedures. The Parallel operator is used to monitor concurrent procedures that are not in conflict by definition. Hence each procedure operates in its own sub-space

A BT-based task planner that makes large use of the Parallel operator is the A Behavior Language (ABL)~\cite{mateas2002behavior}. ABL was originally designed for the dialogue game \emph{Fa\c{c}ade}, and it is appreciated for
its ability to handle planning and acting on multiple scales as, in particular, Real-Time Strategy games~\cite{weber2010reactive}.
ABL executes sub-plans in parallel and  resolves conflicts between multiple concurrent actions by defining a fixed priority order. This jeopardizes the modularity of BTs and introduces the threat of starvation (i.e. the execution of an action with low priority is perpetually denied).

To conclude, there is currently no work addressing the synchronization problems of the Parallel operator. This makes our paper fundamentally different than the ones presented above and the BT literature.

\section{Background:Behavior Trees and Concurrent Programming}
\label{sec:background}

In this section, we briefly describe the semantic of classical BTs and the standard semantics for concurrent programming. A more detailed description of BTs can be found in~\cite{BTBook} while a more detailed description of concurrent programming can be found in~\cite{taubenfeld2006synchronization}.
\subsection{Behavior Trees}
\label{sec:background.BT}

A BT is a graphical modeling language used as a representation for execution of observation-based actions in a system. A BT is represented as a rooted tree where the internal nodes represent operators and leaf nodes represent actuation or sensing skills. 

Graphically, the children of a node are placed below it and they are executed in the order from left to right, as will be explained later.

The execution of a BT begins from the root node. It sends activation signals called \emph{Ticks} with a given frequency to its children. A node in the tree is executed if and only if it receives Ticks from its parent. When the node no longer receives Ticks, its execution is aborted.  The child returns to the parent a status \emph{Success}, \emph{Running}, or \emph{Failure} according to the node's logic. Below we present the most common BT nodes.


When a Fallback operator receives Ticks, it routes them to its own children from the left, returning Success/Running as soon as it finds a child that returns Success/Running. It returns Failure when all the children return Failure. When a child returns Running or Success, the Fallback operator does not send Ticks the next child (if any). 
The Fallback operator is graphically represented by a box with the label ``$?$", as in Figure~\ref{EV.fig.BTEx1}, and its pseudocode is described in Algorithm~\ref{bts:alg:fallback} 

When a Sequence operator receives Ticks, it routes them to its own children from the left, returning Failure/Running as soon as it finds a child that returns Failure/Running. It returns Success when all the children return Success. When a child returns Running or Failure, the Sequence node does not send Ticks the next child (if any).
The Sequence operator is graphically represented by a box with the label ``$\rightarrow$", as in Figure~\ref{EV.fig.BTEx1}, and its pseudocode is described in Algorithm~\ref{bts:alg:sequence} 

The Parallel operator ticks its children in parallel and returns Success if all children return Success, it returns Failure if at least one child returns Failure, and it returns Running otherwise.
The Parallel operator is graphically represented by a box with the label ``$\rightrightarrows$", as in Figure~\ref{EV.fig.BTEx1}, and its pseudocode is described in Algorithm~\ref{bts:alg:parallel}.

\begin{remark}
The for loop in Algorithm~\ref{bts:alg:parallel} Line~2 executes the loops in parallel.
\end{remark}

\begin{algorithm2e}[t!]
\SetKwProg{Fn}{Function}{}{}

\Fn{Tick()}
{
  \For{$i \gets 1$ \KwSty{to} $N$}
  {
    \ArgSty{childStatus} $\gets$ \ArgSty{child($i$)}.\FuncSty{Tick()}\\
    \uIf{\ArgSty{childStatus} $=$ \ArgSty{Running}}
    {
      \Return{Running}
    }
    \ElseIf{\ArgSty{childStatus} $=$ \ArgSty{Success}}
    {
      \Return{Success}
    }
  }
  \Return{Failure}
  }
  \caption{Pseudocode of a Fallback operator with $N$ children}
    \label{bts:alg:fallback}
\end{algorithm2e}

\begin{algorithm2e}[t!]
\SetKwProg{Fn}{Function}{}{}

\Fn{Tick()}
{
  \For{$i \gets 1$ \KwSty{to} $N$}
  {
    \ArgSty{childStatus} $\gets$ \ArgSty{child($i$)}.\FuncSty{Tick()}\\
    \uIf{\ArgSty{childStatus} $=$ \ArgSty{Running}}
    {
      \Return{Running}
    }
    \ElseIf{\ArgSty{childStatus} $=$ \ArgSty{Failure}}
    {
      \Return{Failure}
    }
  }
  \Return{Success}
  }
  \caption{Pseudocode of a Sequence operator with $N$ children}
  \label{bts:alg:sequence}
\end{algorithm2e}

\begin{algorithm2e}[t]
\SetKwProg{Fn}{Function}{}{}

\Fn{Tick()}
{
  \ForAll{$i \gets 1$ \KwSty{to} $N$}
  {
    \ArgSty{childStatus}[i] $\gets$ \ArgSty{child($i$)}.\FuncSty{Tick()}\\
    }
    \uIf{$\Sigma_{i: \ArgSty{childStatus}[i]=Success}1 = N$}
    {
      \Return{Success}
    }
    \ElseIf{$\Sigma_{i: \ArgSty{childStatus}[i] =Failure}1 > 0$}
    {
      \Return{Failure}
    
  }\Else{
  \Return{Running}
  }
  }
    \caption{Pseudocode of a Parallel operator with $N$ children}
  \label{bts:alg:parallel}
\end{algorithm2e}

\begin{algorithm2e}[t]
\SetKwProg{Fn}{Function}{}{}

\Fn{Tick()}
{
 \ArgSty{DoAStepOfComputation()} \\
    \uIf{action-succeeded}
    {
      \Return{Success}
    }
    \ElseIf{action-failed}
    {
      \Return{Failure}
    }
    \Else
    {
    \Return{Running}
    }
   }
  \caption{Pseudocode of a BT Action}
  \label{bts:alg:action}
\end{algorithm2e}

\begin{algorithm2e}[h!]
\SetKwProg{Fn}{Function}{}{}

\Fn{Tick()}
{
    \uIf{condition-true}
    {
      \Return{Success}
    }
    \Else
    {
      \Return{Failure}
    }
   }
  \caption{Pseudocode of a BT Condition}
  \label{bts:alg:condition}
\end{algorithm2e}
As long as an Action receives Ticks, it performs some operations. It returns Success to its parent if the operations are completed and Failure if the operations cannot be completed. Otherwise, it returns Running. Whenever a running Action does no longer receive Ticks, its execution is aborted.
An Action is graphically represented by a rectangle, as in Figure~\ref{EV.fig.BTEx1}, and its pseudocode is described in Algorithm~\ref{bts:alg:action}.

Whenever a Condition receives Ticks, it checks if a proposition is satisfied or not, returning Success or Failure accordingly. A Condition is graphically represented by an ellipse, as in Figure~\ref{EV.fig.BTEx1}, and its pseudocode is described in Algorithm~\ref{bts:alg:condition}.

\begin{remark}
Algorithm~\ref{bts:alg:action} does a step of computation at each Tick. This implementation is preferred in BT libraries for computer games as Unreal Engine and PyGame. 
BT libraries for robotics applications prefer an implementation that allows actions to continue as long as they succeed or fails such as YARP-BT or ROS-BT. Note that in this case whenever a running action no longer receives a tick, its execution is aborted. This requires the implementation specific routine for aborting the action safely.
\end{remark}

%
The state space formulation of BTs~\cite{ColledanchiseTRO17} allows us to study BTs from a mathematical standpoint. In that formulation, the Tick is represented by a recursive function call that includes both the return status, the system dynamics, and the system state.

\begin{definition}[Behavior Tree \cite{ColledanchiseTRO17}]
\label{bts.def:BT}
A BT is a three-tuple 
\begin{equation}
 \bt_i=\{f_i,r_i, \Delta t\}, 
\end{equation}
where $i\in \mathbb{N}$ is the index of the tree, $f_i: \mathbb{R}^n \rightarrow  \mathbb{R}^n$ is the right hand side of a difference equation, $\Delta t$ is a time step and 
$r_i: \mathbb{R}^n \rightarrow  \{\mathcal{R},\mathcal{S},\mathcal{F}\}$ is the return status that can be equal to either 
\emph{Running} ($\mathcal{R}$),
\emph{Success} ($\mathcal{S}$), or
\emph{Failure} ($\mathcal{F}$).


Let the Running/Activation region ($R_i$),
Success region ($S_i$) and
Failure region ($F_i$) correspond to a partitioning of the state space,  defined as follows:
\begin{eqnarray}
 R_i&=&\{x: r_i(x)=\mathcal{R} \} \\
 S_i&=&\{x: r_i(x)=\mathcal{S} \} \\
 F_i&=&\{x: r_i(x)=\mathcal{F} \}. 
\end{eqnarray}
Finally,  let $x_k=x(t_k)$ be the system state at time $t_k$, then the execution of a BT $\bt_i$ is a standard ordinary difference equation
\begin{eqnarray}
 x_{k+1}&=&f_i( x_{k}),  \label{bts:eq:executionOfBT}\\
 t_{k+1}&=&t_{k}+\Delta t.
\end{eqnarray}
 \end{definition}
The return status $r_i$ will be used when combining BTs recursively, as explained below.

Having defined a BT using a state space formulation, Sequence, Fallback, and Parallel compositions are defined as follows:

\begin{definition}[Sequence compositions of BTs \cite{ColledanchiseTRO17}]
\label{bts:def.seq}
 Two or more BTs  can be composed into a more complex BT using a Sequence operator,
 $$\bt_0=\mbox{Sequence}(\bt_1,\bt_2).$$ 
 Then $r_0,f_0$ are defined as follows
\begin{eqnarray}
   \mbox{If }x_k\in S_1&& \\
   r_0(x_k) &=&  r_2(x_k) \\
   f_0(x_k) &=&  f_2(x_k) \label{bts:eq:seq1}\\ 
   \mbox{ else }&& \nonumber \\
   r_0(x_k) &=&  r_1(x_k) \\
   f_0(x_k) &=&  f_1(x_k). \label{bts:eq:seq2}
 \end{eqnarray}
\end{definition}

\begin{definition}[Fallback compositions of BTs \cite{ColledanchiseTRO17}]
\label{bts:def.fal}
 Two or more BTs  can be composed into a more complex BT using a Fallback operator,
 $$\bt_0=\mbox{Fallback}(\bt_1,\bt_2).$$ 
 Then $r_0,f_0$ are defined as follows
\begin{eqnarray}
   \mbox{If }x_k\in {F}_1&& \\
   r_0(x_k) &=&  r_2(x_k) \\
   f_0(x_k) &=&  f_2(x_k) \\ 
   \mbox{ else }&&\nonumber \\
   r_0(x_k) &=&  r_1(x_k) \\
   f_0(x_k) &=&  f_1(x_k).
 \end{eqnarray}
\end{definition}

\begin{definition}[Parallel compositions of BTs \cite{ColledanchiseTRO17}]
 \label{bts:def.par}
 Two or more BTs  can be composed into a more complex BT using a Parallel operator,
 $$\bt_0=\mbox{Parallel}(\bt_1,\bt_2).$$ 
 Let $x_k=(\tilde x_{1k}, \tilde x_{2k})$ be a partitioning of the state space such that $\tilde x_{1k} \perp \tilde x_{2k}$,
 then $r_0$, $f_0$ are defined as follows
\begin{eqnarray}
   r_0(x_k) &=& \begin{cases}
   \mathcal{S}&  \mbox{ If } r_1(\tilde x_{1k})=\mathcal{S} \wedge r_2(\tilde x_{2k})=\mathcal{S}\\ 
   \mathcal{F}&  \mbox{ If } r_1(\tilde x_{1k})=\mathcal{F} \vee r_2(\tilde x_{1k})=\mathcal{F}\\   
   \mathcal{R}&  \mbox{ else } 
   \end{cases}\\
      f_0(x_k) &=& (f_{1}(\tilde x_{1k}),f_{2}(\tilde x_{2k}))
 \end{eqnarray}
\end{definition}
%
%
\subsection{Concurrent Programming}

\label{sec:background.cp}

Concurrent programming refers to the execution of multiple processes during overlapping periods of time concurrently instead of sequentially.

The need for synchronization arises in any kind of concurrent process, even in single-processor systems. The main needs for synchronization are: \emph{producer-consumer relationship}, where a consumer process has to wait until a producer produces the necessary data, and \emph{exclusive use of resources}, where multiple processes want to access a critical resource. A synchronization strategy must ensure that only one process at a time can access the resource~\cite{taubenfeld2006synchronization}.


\emph{Barriers} and \emph{Semaphores} are popular ways to implement synchronization.
A Barrier is a method that allows concurrent processes to wait for each other at a specific point of execution. A Semaphore is a method that regulates the access to critical resources.

\emph{Deadlocks} and \emph{Starvation} are among the most common threats in concurrent programming\cite{taubenfeld2006synchronization}. A Deadlock occurs when processes in a group are all waiting for resources that are being held by the others. In this case, the processes just keep waiting and execute no further;
There are several techniques to avoid deadlocks~\cite{taubenfeld2006synchronization}.
A Starvation occurs when a process is waiting to use shared resource, but other processes monopolize it (e.g. because they have higher priority), and the first process is forced to wait indefinitely. A solution to starvation is to implement \emph{aging}, where all process are preemptable and a process waiting a long time to use a resource gradually increases its chances to use it.

\section{Problem Formulation}
\label{sec:problem}
In this section, we first make a set of assumptions and definitions, then state the main problems.

To be able to synchronize the execution of concurrent BTs, they need to indicate their progress and the resources used.
Hence we make the following assumptions.

\begin{assumption}
\label{pf.ass.progress}
For each BT node there exists a function that indicates the progress of the node's execution at each state.
\end{assumption}


%


\begin{assumption}
\label{pf.ass.resources}
For each BT node there exists a function that indicates the resources used in its execution.
\end{assumption}

To guarantee that only the ticked actions are executed, whenever a running action no longer receives ticks it stops its execution within a negligible time with respect to the time step $\Delta t$ of Definiton~\ref{bts.def:BT}. Hence we make the following assumption.

\begin{assumption}
\label{pf.ass.halt}
Each action is able to stop its execution within $\tau << \Delta t$ when it no longer receives ticks. During  $\tau$ the node does not change its progress value or resources needed.  
\end{assumption}

%


We allow disturbance to affect the progress value. However, we assume that the disturbance acts only on nodes while they are running. Hence we make the following assumption.

\begin{assumption}
\label{pf.ass.progressrunning}
A BT can change its progress only while it is running.
\end{assumption}

For convergence, we need to state the following assumptions.

\begin{assumption}
\label{pf.ass.finite}
An action node will terminate its execution in finite time.
\end{assumption}

\begin{definition}[Progress independent CBTs]
\label{pf.def.progressindepended}
Two CBTs, $\bt_1$ and $\bt_2$, are said progress independent if the progress of a BT does not depend on the progress of the other. 
\end{definition}

We are now ready to formulate the main problems:

\begin{problem}
\label{problem.sync}
Let $\bt_1$ and $\bt_2$ be two progress independent BTs. Execute the two BTs concurrently such that $\bt_1$ and $\bt_2$ are both running only when they have the same progress, otherwise the BT with the highest progress waits.
\end{problem}
\begin{remark}
$\bt_1$ and $\bt_2$ needs to be progress independent to avoid conflicts with the progress dependence that will be introduced.
\end{remark}

\begin{problem}
\label{problem.mutex}
Let $\bt_1$ and $\bt_2$ be two BTs. 
Execute the two BTs concurrently in a deadlock-free and starvation-free fashion such that they do not use the same resources when they are running.
\end{problem}

\begin{remark}
Problem~\ref{problem.mutex} relaxes the assumption that two CBTs operate on orthogonal state spaces, as stated in Definition~\ref{bts:def.par}.
\end{remark}

\section{Proposed Solution}
\label{sec:proposed}
In this section, we present the proposed solution to Problems~\ref{problem.sync} and~\ref{problem.mutex}. 
We first define the progress and the resource function, needed to formally define CBTs. We then propose two new parallel operators to solve Problems~\ref{problem.sync} and~\ref{problem.mutex}.
We define the operators by theirs pseudocode and their space state formulation, following the formulation of~\cite{ColledanchiseTRO17}. 
\begin{definition}[Progress Function]
\label{ps.def.progress}
The function $p: R^n \to [0,1]$ is the progress function. It indicates the progress of the BT's execution at each state.
\end{definition}

\begin{definition}[Resources]
\label{ps.def.L}
$L$ is a collection of symbols that represents the resources available in the system.
\end{definition}

\begin{definition}[Resource Function]
\label{ps.def.resources}
The function \\ $q : R^n \to 2 ^ L$ is the resource function. It indicates the set of resources needed for a BT's execution at each state.
\end{definition}

We can now define a CBT as BT with information regarding its progress and the resources needed as follows:

\begin{definition}[Concurrent BTs]
\label{bts.def:CBT}
A CBT is a tuple 
\begin{equation}
 \bt_i=\{f_i,r_i, \Delta t, p_i, q_i\}, 
\end{equation}
where $i$, $f_i$, $\Delta t$, 
$r_i$ are defined as in Definition~\ref{bts.def:BT}, $p_i$ is a progress function, and  $q$ is the resource function.
\end{definition}

A CBT has the functions $p_i$ and $q_i$ in addition to the others of Definition~\ref{bts.def:BT}. These functions are user-defined for Actions and Condition. For the classical operators, the functions are defined below.

\begin{definition}[Sequence compositions of CBTs]
\label{bts:def.smoothseq}
 Two CBTs  can be composed into a more complex CBT using a Sequence operator,
 $$\bt_0=\mbox{Sequence}(\bt_1,\bt_2).$$ 
 The functions $r_0,f_0$ match those introduced in Definition~\ref{bts:def.seq}, while the functions $p_0,q_0$ are defined as follows
\begin{eqnarray}
   \mbox{If }x_k\in S_1&& \\
   p_0(x_k) &=& 0.5 + \frac{p_2(x_k)}{2} \\
   q_0(x_k) &=& q_2(x_k)\\
   \mbox{ else }&& \nonumber \\
   p_0(x_k) &=& \frac{p_1(x_k)}{2}\\
   q_0(x_k) &=& q_1(x_k).
 \end{eqnarray}
\end{definition}

\begin{definition}[Fallback compositions of CBTs]
\label{bts:def.smoothfal}
 Two CBTs  can be composed into a more complex CBT using a Fallback operator,
 $$\bt_0=\mbox{Fallback}(\bt_1,\bt_2).$$ 
 The functions $r_0,f_0$ are defined as in Definition~\ref{bts:def.fal}, while the functions $p_0,q_0$ are defined as follows
\begin{eqnarray}
   \mbox{If }x_k\in {F}_1&& \\
   p_0(x_k) &=& p_2(x_k)\\
   q_0(x_k) &=& q_2(x_k) \\ 
   \mbox{ else }&&\nonumber \\
   p_0(x_k) &=& p_1(x_k)\\
   q_0(x_k) &=& q_1(x_k).
 \end{eqnarray}
\end{definition}

\begin{definition}[Parallel compositions of CBTs]
 \label{bts:def:smoothpar}
 Two CBTs  can be composed into a more complex CBT using a Parallel operator,
 $$\bt_0=\mbox{Parallel}(\bt_1,\bt_2).$$ 
 The functions $r_0,f_0$ are defined as in Definition~\ref{bts:def.par}, while the functions $p_0$ and $q_0$ are defined as follows
\begin{eqnarray}
   p_0(x_k) &=& \mbox{min}(p_1(x_k), p_2(x_k)) \\ 
   q_0(x_k) &=& q_1(x_k) \cup q_2(x_k) \\
 \end{eqnarray}
\end{definition}
\begin{remark}
Conditions nodes do not perform any action. Hence their progress function can be defined as $p(x_k) = 1$. 
\end{remark}
\subsection{Process Synchronization of CBTs}
\label{sec.smooth.parallel}
In this section, we show how by using CBTs we can synchronize the progress of two or more actions executed in parallel, solving Problem~\ref{problem.sync}.

In the classical formulation, the Parallel operator (see Definition~\ref{bts:def.par}) requires that the BTs are not dependent on each other, hence it is not possible to synchronize their execution.
Having extended the BTs formulation with the progress function, we can define a  Synchronized Parallel operator that takes into account the BTs progress when deciding where to route the Ticks. 

\begin{definition}[Synchronized Parallel]
\label{PS:def.ParallelSync}
 Two CBTs  can be composed into a more complex BT using a Synchronized Parallel operator,
 $$\bt_0=\mbox{ParallelSync}(\bt_1,\bt_2).$$ 
 
Let $x_k=(\tilde x_{1k}, \tilde x_{2k})$ be the partitioning  as in Definition~\ref{bts:def.par}
 then $f_0 $ is defined as follows:
\begin{eqnarray}
	f_0 (x_k) &=& (\tilde f_1(\tilde x_{1k}), \tilde f_2(\tilde x_{2k}))\\
	   \tilde f_i(x_k)  &=& \begin{cases} f_i(\tilde x_i) &\mbox{if }  \alpha_i(x_k)  = 1  \\ \tilde x_i &\mbox{otherwise} \end{cases}  \\ 
   \alpha_1(x_k)  &=& \begin{cases} 0 &\mbox{if } p_1(\tilde x_{1k}) > p_2(\tilde x_{2k})\\ 1 &\mbox{otherwise} \end{cases}  \\ 
   \alpha_2(x_k)  &=& \begin{cases} 0 &\mbox{if } p_2(\tilde x_{2k}) > p_1(\tilde x_{1k})\\ 1 &\mbox{otherwise} \end{cases}   
 \end{eqnarray}
\end{definition}

%
%


Algorithm~\ref{ps.alg.synch} presents the pseudocode of a Synchronized Parallel operator.
The main difference with the classical Parallel operator (Algorithm~\ref{bts:alg:parallel}) lies in the addition of Lines~2-4 and Line~6, which implement the fact that a child receives Ticks only if its progress does not exceed the minimum progress. We graphically represent this operator with box with the label ``$\overset{S}{\rightrightarrows}$".

\begin{algorithm2e}[h]
\SetKwProg{Fn}{Function}{}{}

\Fn{Tick()}
{
\ArgSty{minProgress} $\gets$ 1 \\ 
  \For{$i \gets 1$ \KwSty{to} $N$}
  {
  \ArgSty{minProgress} $\gets$ \FuncSty{min}(\ArgSty{minProgress}, $p_i$)
  }
  \ForAll{$i \gets 1$ \KwSty{to} $N$}
  {

  \If{$p_i \leq$ \ArgSty{minProgrees}}{
    \ArgSty{childStatus}[i] $\gets$ \ArgSty{child($i$)}.\FuncSty{Tick()}\\
  }
    }
    \uIf{$\Sigma_{i: \ArgSty{childStatus}[i]=Success}1 = N $}
    {
      \Return{Success}
    }
    \ElseIf{$\Sigma_{i: \ArgSty{childStatus}[i]=Failure}1 > 0$}
    {
      \Return{Failure}
    
  }
  \Return{Running}
  
  }
    \caption{Pseudocode of a ParallelSync node with $N$ children}
 \label{ps.alg.synch}
\end{algorithm2e}

\begin{remark}
The Synchronized Parallel operator intrinsically implements  a barrier (as described in Section~\ref{sec:background.cp}) at the minimum progress value, where all children wait for the one that has the ``slowest" value to proceed. 
\end{remark}

We are now ready the see the execution of the BT in Figure~\ref{intro.fig.sync} formalized in the example below.
\begin{example}
\label{PS.example.seek}
An object-seeking robot has to recognize objects on the floor. The robot's behavior is described by the BT in Figure~\ref{intro.fig.sync}. The progress of both actions is $0$ at the beginning of the hallway and $1$ when the hallway is completely navigated or sought, it is a value in $(0,1)$ according to the percentage of the hallway navigated or swept. If the robot finds an object in the middle of the hallway where progress for both actions is $0.5$. Since it takes time to compute the perception routine that allows the robot to recognize the object, the progress of \emph{Seek} is equal to $0.5$ until the robot recognizes the object. The navigation action could continue its progress. However, as soon as its progress surpasses the value $0.5$ the navigation no longer receives Ticks (Algorithm~\ref{ps.alg.synch} Lines~6-7) and the execution is aborted, making the navigation stop. Once the robot recognizes the object, the progress of \emph{Seek} increases. As soon as the progress of \emph{Navigate} is equal to the progress of \emph{Seek} the navigation resumes. 
\end{example}

\subsection{Resource Synchronization of CBTs}
In this section we show how using CBTs we can execute multiple actions in parallel without resource conflicts.


\begin{definition}[Mutually Exclusive Parallel]
\label{ps.def.mutex}
 Two BTs  can be composed into a more complex BT using a Mutually Exclusive Parallel operator,
 $$\bt_0=\mbox{ParallelMutex}(\bt_1,\bt_2).$$ 
$f_0$ and $q_0$ are defined as follows:
\begin{eqnarray}
	f_0(x_k) &=& f_1(x_k)\alpha_1(x_k) +  f_2(x_k)\alpha_2(x_k) \\ 
	q_0(x_k) &=& \bigcup_{i: \alpha_i(x_k) = 1} q_i(x_k)\\
   \alpha_1(x_k) &=& \begin{cases} 1 &\mbox{if } q_1(x_k) \cap  q_2(x_k) = \emptyset \; \lor \\ & \hspace{1em}  \lor \pi_1(x_k) \geq \pi_2(x_k) \\ 0 &\mbox{otherwise} \end{cases}  \\
      \alpha_2(x_k) &=& \begin{cases} 1 &\mbox{if } q_1(x_k) \cap  q_2(x_k) = \emptyset \; \lor \\ & \hspace{1em}  \lor \pi_2(x_k)  > \pi_1(x_k) \\ 0 &\mbox{otherwise} \end{cases} \\   
 \end{eqnarray}
\end{definition}

where $\pi_i(x_k)$ a user-defined priority assigned by any policy that implements \emph{aging} (see Section~\ref{sec:background.cp}).

Algorithm~\ref{ps.alg.mutex} presents the pseudocode of a Mutually Exlusive Parallel operator.
The main difference with the classical Parallel operator (Algorithm~\ref{bts:alg:parallel}) lies in Lines~1-7 and Line~9, which implement the fact that a child nodes receives ticks only if it uses a resource that is not required by any currently running node. We graphically represent the operator with a box with the label ``$\overset{M}{\rightrightarrows}$".

\begin{remark}
As mentioned in Section~\ref{sec:background.cp}, using a policy that implements aging avoids starvation.
\end{remark}

\begin{algorithm2e}[t]
\ArgSty{$\tilde R$} $\gets \emptyset$  \\ 
\ArgSty{maxPrority} $\gets 0$\\
  \For{$i \gets 1$ \KwSty{to} $N$}
  {
  \ArgSty{maxPrority} $\gets$ \FuncSty{max}(\ArgSty{maxPrority}, $\pi_i$)
  }
  \ForAll{$i \gets 1$ \KwSty{to} $N$} 
  {
  \If{$q_i \not\subseteq$ \ArgSty{ $\tilde R$} \textbf{or} $\pi_i = maxPrority$}{
      \ArgSty{$\tilde R$} $\gets$ \ArgSty{$\tilde R \cup q_i$} \\ 
    \ArgSty{childStatus}[i] $\gets$ \ArgSty{Children($i$).\FuncSty{Tick()}}\\
       } \ArgSty{$\tilde R$} $\gets \emptyset$\\
    }
    \uIf{$\Sigma_{i: \ArgSty{childStatus}[i]=Success}1\geq N$}
    {
      \Return{Success}
    }
    \ElseIf{$\Sigma_{i: \ArgSty{childStatus}[i]=Failure}1 > 0$}
    {
      \Return{Failure}
  }
  \Return{Running}
    \caption{Pseudocode of a ParallelMutex node with $N$ children}
\label{ps.alg.mutex}
\end{algorithm2e}

We now ready the describe the execution of the BT in Figure~\ref{intro.fig.mutex}, formalized in the example below.

\begin{example}
\label{PS.example.speech}
A delivery robot is asked to distribute mail to an office's employee. During its execution, the robot monitors the possible position of people around to avoid harming them. The robot's behavior is described by the BT in Figure\ref{intro.fig.mutex}. If a person is obstructing any obstacle-free path, it asks the person to move away. Moreover, if the room is too dark it asks to improve the light condition, for better robot's perception.
Without loss of generality, let's assume that a person gets in front of the robot before this realizes that the room is too dark. ParallelMutex has $\tilde R = \emptyset$ (Algorithm~\ref{ps.alg.mutex} Line~1) and it ticks the action \emph{Monitor People} first. Hence its set $q(x_k)$ (set of resource used) contains the symbol \emph{speaker} making $\tilde R$ also contain the symbol \emph{speaker} (Algorithm~\ref{ps.alg.mutex} Line~5). When the robot realizes that the room is too dark, the action \emph{Monitor Light} is not ticked as its set $q(x_k)$ contains the symbol \emph{speaker} and the condition in Algorithm~\ref{ps.alg.mutex} Line~4 is not satisfied (i.e. the speaker is not available at that time).
\end{example}
%
%
%
%

\section{Theoretical Analysis}
\label{sec:analysis}
In this section, we prove that the proposed approach solves Problems~\ref{problem.sync} and~\ref{problem.mutex}.

\begin{proposition}
Let $\bt_1$ and $\bt_2$ be two progress independent BTs, $\bt_0 = \mbox{ParallelSync}(\bt_1,\bt_2)$ solves Problem~\ref{problem.sync}.
\begin{proof}
The execution of $\bt_0$ is defined by Algorithm~\ref{ps.alg.synch}. 
For each state $x_k \in R^n$, one of the following occurs:
\begin{itemize}
\item $p_1(x_k) = p_2(x_k)$: in this case both BTs are executed as they both receive Ticks.
\item $p_1(x_k) < p_2(x_k)$: Since 
in this case \emph{minProgress} (Algorithm~6 Line~4) is $p_1(x_k)$ and $\bt_2$ no longer receive Ticks (Line~6) making the execution of $\bt_2$ stop, hence $p_2(x_{k+1}) = p_2(x_k)$. 
\item $p_2(x_k) < p_1(x_k)$: $p_1(x_{k+1}) = p_1(x_k)$ holds similarly as above. 
\end{itemize}
\end{proof}
\end{proposition}
Moreover, by Assumptions~\ref{pf.ass.halt} and~\ref{pf.ass.progressrunning} a BT stops increasing or decreasing its progress whenever it no longer receives ticks. Hence, $\bt_1$ and $\bt_2$ are both running only if they have the same progress, otherwise the BT with the higher progress waits for the other.

\begin{lemma}
Let $\bt_1$ and $\bt_2$ be two progress independent BTs, the execution of $\bt_0 = \mbox{ParallelMutex}(\bt_1,\bt_2)$ is deadlock-free.

\begin{proof}
By Assumption~\ref{pf.ass.finite} all the actions either terminate or set the value of the resource function to the empty set. Hence the resources are released in finite time.
\end{proof}
\end{lemma}

\begin{lemma}
Let $\bt_1$ and $\bt_2$ be two progress independent BTs, the execution of $\bt_0 = \mbox{ParallelMutex}(\bt_1,\bt_2)$ is starvation-free.

\begin{proof}
In Algorithm~\ref{ps.alg.mutex} assign resources using a policy that implements aging.
\end{proof}
\end{lemma}

\begin{proposition}
Let $\bt_1$ and $\bt_2$ be two progress independent BTs, $\bt_0 = \mbox{ParallelMutex}(\bt_1,\bt_2)$ solves Problem~\ref{problem.mutex}.
\begin{proof}
The execution of $\bt_0$ is defined by Algorithm~\ref{ps.alg.mutex}. 
For each state $x_k \in R^n$, one of the following occur:
\begin{itemize}
\item $q_1(x_k) \cup q_2(x_k) = \emptyset$ : in this case the condition in Algorithm~7 Line~2, trivially holds for both children, which are ticked.
\item $q_1(x_k) \cup q_2(x_k) \neq \emptyset$: Only the child with highest priority is executed.
\end{itemize}
hence $\bt_1$ and $\bt_2$ do not use the same resources when they are running. By Lemma~1 and Lemma~2 the execution of $\bt_0$ is deadlock-free and starvation-free.
\end{proof} 
\end{proposition}

\section{Implementation Details}
\label{sec:implementation}
To allow CBTs to be implemented, we must make some key observations.

\subsection{Wait and Halt Routines}

In the formulation above, a running action no longer receives Ticks in one of these two cases: when the parent is routing the Ticks elsewhere (i.e. it no longer requires the execution of that action) and when the action has to wait for another one to reach the progress barrier (in case of Synchronized Parallel) or some resources (in case of Mutually Exclusive Parallel). 
As mentioned in Section~\ref{sec:background}, an action node may require to execute a specific routine whenever the nodes no longer receive ticks. It is fundamental to define two different routines, one when the action is no longer required (\emph{halt}) and one when the action has to wait for another one (\emph{pause}) (e.g. a walking action moves the robot on a stable position with both feet on the ground when it has to stop, whereas it slows down speed of the motors when waiting for another action). Algorithm~\ref{impl:alg:parallel} shows the pseudocode of a Synchronized Parallel operator that explicitly defines these two routines. The Mutually Exclusive Parallel operator can be extended similarly.

\begin{algorithm2e}[t]
\SetKwProg{Fn}{Function}{}{}

\Fn{Tick()}{

\ArgSty{minProgrees} $\gets$ 1 \\ 
  \For{$i \gets 1$ \KwSty{to} $N$}
  {
  \uIf{$p_i \leq$ \ArgSty{minProgrees}}
  {
    \ArgSty{childStatus[i]} $\gets$   \ArgSty{children[i]}.\FuncSty{Tick()}\\
    \ArgSty{minProgrees} $\gets$ \FuncSty{min}(\ArgSty{minProgrees}, $p_i$)}
    }\Else
    {
    \ArgSty{children[i]}.\FuncSty{Pause()}\\
    }

    \uIf{$\Sigma_{i: \ArgSty{childStatus}[i]=Success}1\geq M$}
    {
      \Return{Success}
    }
    \ElseIf{$\Sigma_{i: \ArgSty{childStatus}[i]=Failure}1 > N-M$}
    {
      \For{$i \gets 1$ \KwSty{to} $N$}
  {
      \ArgSty{children[i]}.\FuncSty{Halt()}\\
  }
      \Return{Failure}
    
  }
  \Return{Running}
  }
  \Fn{Halt()}
  {
    \For{$i \gets 1$ \KwSty{to} $N$}
  { 
  children[i].\FuncSty{Halt()}
  }
  }
  
    \Fn{Pause()}
  {
    \For{$i \gets 1$ \KwSty{to} $N$}
  { 
  children[i].\FuncSty{Pause()}
  }
  }
    \caption{Pseudocode of a ParallelSync node with $N$ children with explicit handling of the Halt and Pause routine.}
  \label{impl:alg:parallel}
\end{algorithm2e}

%

\subsection{Resolution of Progress Function}
It sometimes impossible to define the progress value of a BT Action with a closed-form expression. This could result in the manual definition of the progress with a coarse resolution, with fixed value on a large portion of the state space. This can jeopardize the synchronized execution of two CBTs (e.g. the extreme case is where an action sets its progress to $1$ only when it is completed and $0$ otherwise).

\subsection{Resource Assignment Policy}
In Definition~\ref{ps.def.mutex} and Algorithm~\ref{ps.alg.mutex}, the user has to define a policy to access a critical resource. The policy has to implement aging to avoid starvation while it must give the resources to the current running action for enough time (e.g. a speech action must have a high chance to use the speaker while the robot is speaking). In a modular framework as BTs the choice of the priority value can require effort, since in some cases we must ensure that the priority of a node is the highest one. However, task scheduling algorithms provide effective solutions.


\section{Use Cases Examples}
\label{sec:results}
In this section, we show two use cases in realistic scenarios where the problems formulated apply. One case involves the Syncronized Parallel operator and the other involves the Mutually Exclusive Parallel operator. 

\begin{figure}[h]
\centering
\includegraphics[width=\columnwidth]{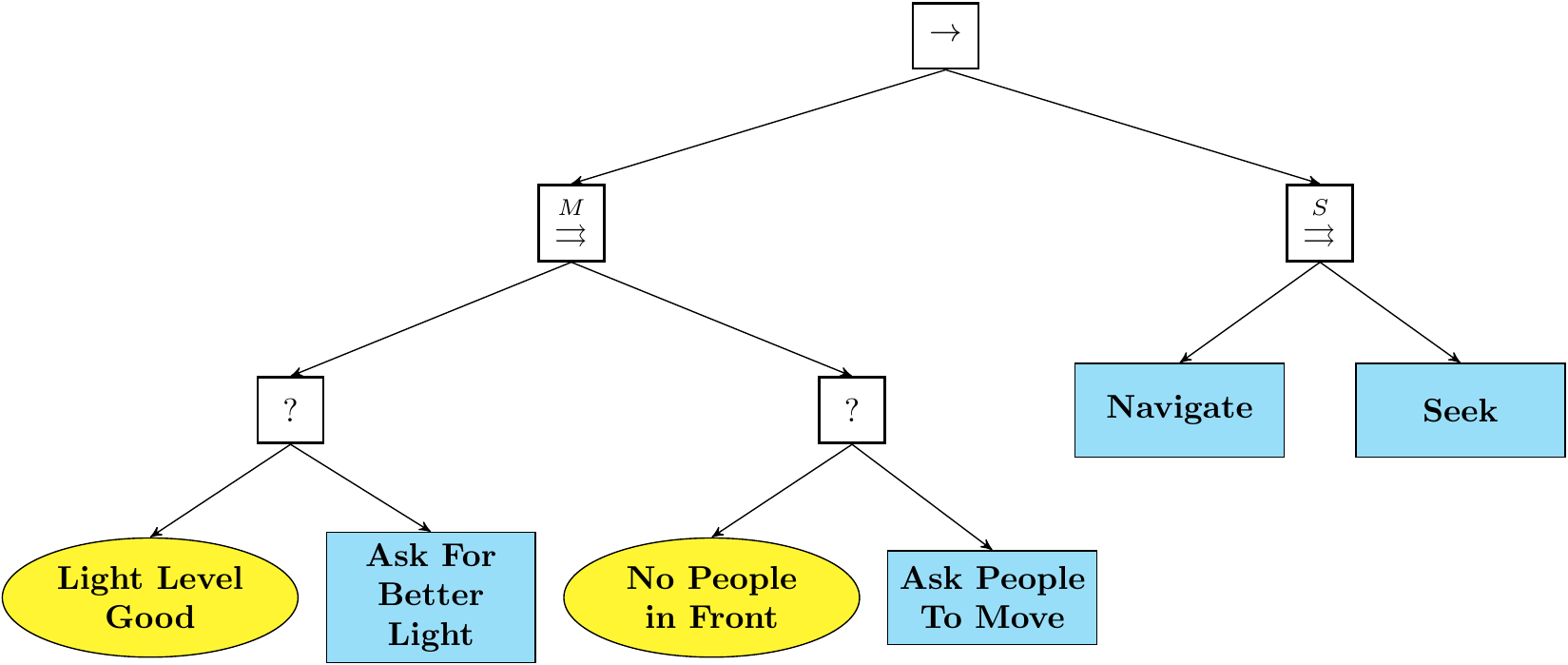}
\caption{BT of Cases~\ref{EV.experiment.seek} and~\ref{EV.experiment.speech}. }
\label{EV.fig.BTEx1}
\end{figure}

In both cases we use the BT in Figure~\ref{EV.fig.BTEx1}.
The action \emph{Navigate} makes the robot follow a collision-free path, If such path exists. The action \emph{Ask People to Move} uses the speaker to ask to make way. The action \emph{Seek} makes the robot move its head looking for objects. Whenever an object is found, the robot stops moving the head until the object is correctly recognized.

\begin{experiment}[Action Synchronization]
\label{EV.experiment.seek}
This case is an implementation of Example~\ref{PS.example.seek}.
When the BT starts its execution, both actions are executed and their progress advances as expected. When the robot finds an object on the floor, it takes time to recognized the object. This blocks the progress of the action \emph{Sweep}. As a consequence, the robot stops navigating because the action \emph{Navigate} no longer receive Ticks. When the object is recognized, the progress of Seek restarts, making the robot resume its navigation. Similarly, when the robot has to avoid obstacles, the Navigate action progresses slowly, as a consequence, the progress of the action \emph{Seek} slows down. 
\end{experiment}

\begin{experiment}[Exclusive Use of Resources]
\label{EV.experiment.speech}
This case is an implementation of Example~\ref{PS.example.speech}. To reach an employee's desk, the robot enters a dark office. In this situation the condition \emph{Light Level Good} returns false, making the ticks reach the action \emph{Ask For Better Light}, which asks to turn on lights four times. An employee hears the robot's request and moves to the switch to turn on the lights. While doing so the employee moves in front of the robot obstructing its path. The robot is now unable to find another path and the condition \emph{No People in Front} returns Failure. In this case, the action \emph{Ask People to Move} cannot be executed as it requires the speaker to be available but it is used by the action \emph{Ask For Better Light}. After a while the situation remains unchanged, the room is still dark and the employee is still in front of the robot. The priority of the action \emph{Ask People to Move} becomes the highest one. The BT stops sending ticks to \emph{Ask For Better Light} and starts sending ticks to \emph{Ask People to Move}.
\end{experiment}

In both use cases the environment can be dynamic and unpredictable, a case where BTs shows advantages over other classical control architecture~\cite{BTBook}. 

\section{Conclusions}
\label{sec:conclusions}
In this paper, we proposed CBTs a generalization of BTs that allows the definition of two new control flow nodes for data and progress synchronization. We defined CBTs and the two new operators according to the state space formulation of BTs, in line with~\cite{BTBook}, analyzed the approach from a theoretical standpoint and provided use cases. This is the first step towards the synchronization of concurrent BTs.
 
\bibliographystyle{IEEEtran}
\bibliography{behaviorTreeRefs}
\end{document}